%% file: main.tex
\pdfoutput=1
\documentclass[letterpaper,10pt,journal,twoside]{IEEEtran}

\usepackage{microtype}
\usepackage{amsmath}
\usepackage{amssymb}
\usepackage{booktabs}
\usepackage{graphicx}
\usepackage{cite}
\usepackage[caption=false,font=footnotesize]{subfig}
\usepackage{hyperref}
\usepackage{xcolor}

\hypersetup{
    colorlinks=true,
    linkcolor=blue,
    urlcolor=blue,
    citecolor=blue,
    pdftitle={Action-Prior Denoising for Smooth Real-Time Chunking},
    pdfauthor={Dongyang Liu, Zhaowen Zheng, Yu Sun, Longxu Zhang, Yixuan Liu, and Hao Wan}
}

\title{Action-Prior Denoising for Smooth Real-Time Chunking}
\author{Dongyang Liu, Zhaowen Zheng, Yu Sun, Longxu Zhang, Yixuan Liu, and Hao Wan%
\thanks{Dongyang Liu and Zhaowen Zheng contributed equally.}%
\thanks{Corresponding author: Hao Wan (wanhao@rokae.com).}%
\thanks{Dongyang Liu, Zhaowen Zheng, and Hao Wan are with ROKAE (Shandong) Robot Group Co., Ltd. Emails: 1721316248@qq.com, zhengzhaowen0918@163.com, wanhao@rokae.com.}%
\thanks{Yu Sun and Longxu Zhang are with the School of Mathematical Sciences, University of Chinese Academy of Sciences, Beijing, China. Emails: 778516650@qq.com, 1970436781@qq.com.}%
\thanks{Yixuan Liu is with The Hong Kong University of Science and Technology (Guangzhou), Guangzhou, China. Email: 1715743372@qq.com.}}
\date{}

\IEEEaftertitletext{%
\vspace{-0.5\baselineskip}
\noindent\begin{minipage}{\textwidth}
    \centering
    \includegraphics[width=0.96\textwidth]{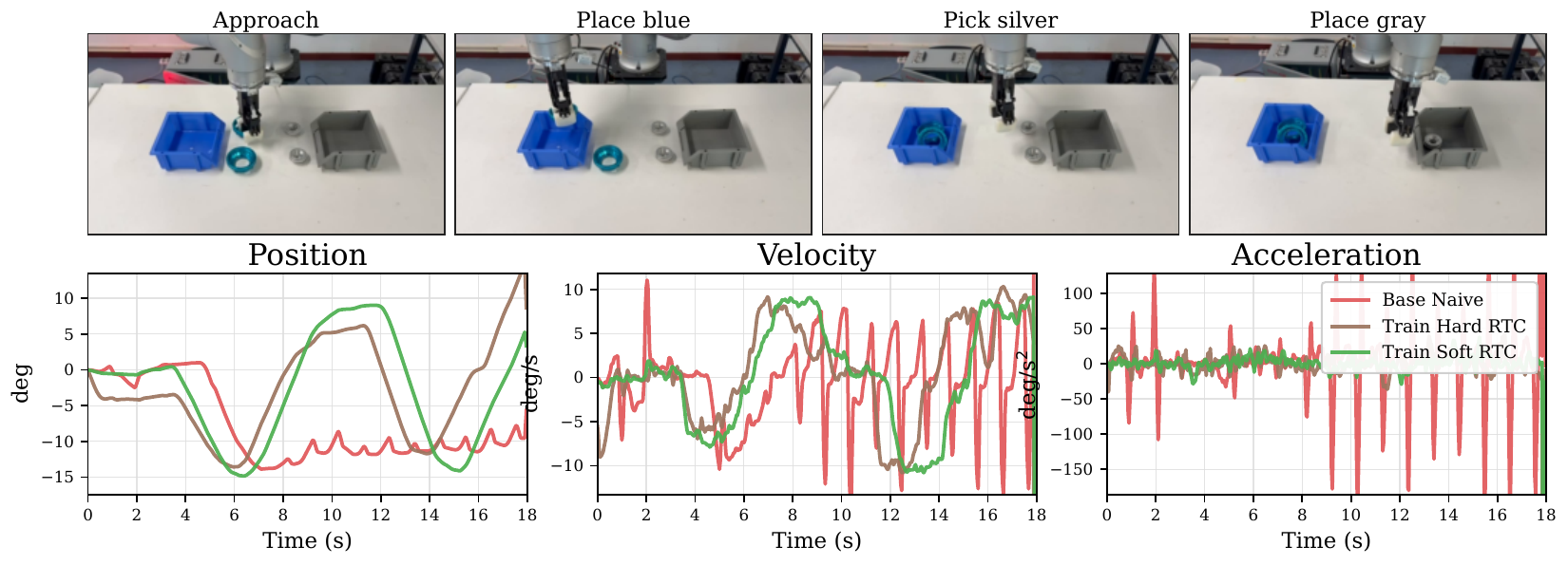}\\[-0.2ex]
    \refstepcounter{figure}\label{fig:teaser}%
    \parbox{0.96\textwidth}{\footnotesize
    \textbf{Fig.~\thefigure.} Teaser. Top: representative frames from a Soft RTC real-robot sorting rollout. Bottom: commanded joint traces for Base Naive, Train Hard RTC, and Train Soft RTC. The plotted velocity and acceleration traces are lightly smoothed for visualization; quantitative command-continuity metrics are reported in Table~\ref{tab:real-robot}.}
\end{minipage}
\vspace{0.4\baselineskip}
}

\begin{document}
\maketitle

\begin{abstract}
Real-time chunking (RTC) lets chunked action policies operate under inference delay by conditioning a newly generated action chunk on actions already committed by the previous chunk. Training-time RTC simulates this delay during learning and avoids expensive guidance at deployment, but its binary prefix mask treats all non-prefix tokens as fully unconstrained. This under-models asynchronous execution: early overlap actions are fixed, while later overlap actions remain editable but should still stay close to the previous plan. We propose \emph{Soft RTC}, a training-time RTC generalization based on action-prior denoising. Soft RTC constructs corrupted overlap tokens from partially denoised states instead of pure noise and injects the aligned previous chunk as the same prior during inference through a lightweight token-wise blending rule. On the 12 released large Kinetix levels, a short soft window nearly matches hard training-time RTC in overall solve rate ($0.809$ vs.\ $0.815$), while a medium window reduces high-delay action delta and jerk by $9.1\%$ and $9.6\%$ relative to hard RTC. Both variants keep near-naive runtime, unlike inference-time RTC baselines. A small preliminary real-robot sorting study provides additional evidence that training-time RTC can improve completion and that Soft RTC gives the lowest commanded-action finite-difference metrics among the tested policies.
\end{abstract}

\begin{IEEEkeywords}
Robot learning, Learning from demonstration, Machine learning for robot control, Motion control, Real-time systems.
\end{IEEEkeywords}

\section{Introduction}

Action chunking has become a common strategy for learned robot control because it amortizes expensive model inference over multiple future actions. Instead of predicting a single action at every control step, a policy predicts a short action sequence and executes only an initial prefix before replanning. This improves temporal consistency and allows larger generative policies to run at meaningful control rates. Action chunking is now central to both transformer-based imitation policies and generative visuomotor policies \cite{zhao2023act,shafiullah2022bet,lee2024vqbet,chi2023diffusion,haldar2024baku}.

However, chunking does not eliminate the real-time execution problem. If model inference takes nontrivial wall-clock time, some actions from the next chunk must still be executed before that chunk finishes generating. Naively waiting causes pauses; naively switching chunks causes abrupt out-of-distribution transitions because the new chunk was generated from a stale observation and is not aligned with actions that have already been committed. On hardware, this can appear as back-and-forth joint motion, local stalling near contact or placement regions, and delayed task completion. Real-Time Chunking (RTC) addresses this by overlapping generation with execution and conditioning the next chunk on the previously committed actions \cite{black2025rtc}. Inference-time RTC is effective, but its inpainting-style conditioning introduces extra compute exactly where latency is already scarce.

Training-time RTC offers a deployment-efficient alternative: simulate inference delay during training and condition on the committed prefix directly \cite{black2025train}. This preserves the efficiency of ordinary policy execution because deployment remains a lightweight forward policy call. Yet the training objective remains binary. It distinguishes between a fully fixed prefix and a fully free suffix, even though real asynchronous chunk execution contains a richer overlap geometry: the earliest overlap actions are irrevocably committed, later overlap actions are still editable but should remain close to the previous chunk, and only the non-overlap tail is fully unconstrained.

This observation motivates the central question of this paper: can the binary prefix conditioning of training-time RTC be generalized into a continuous action-prior denoising objective, without giving up the fast deployment profile of training-time RTC?

From a generative-model perspective, this suggests action-prior denoising. Once a token lies inside the overlap, generating it as if it were unconstrained noise ignores information already available from the previous chunk. A smoother delayed controller should instead denoise from this prior when it exists, and relax the prior as the token moves away from the committed prefix.

We answer this question with \emph{Soft RTC}. The method replaces binary prefix conditioning with a token-wise overlap profile. Early overlap tokens are seen by the model as nearly fixed, later overlap tokens denoise from partially prior-informed states, and free postfix tokens follow ordinary flow matching. The resulting objective requires no architectural changes and contains hard training-time RTC as a special case.

Our contributions are:
\begin{enumerate}
    \item a formulation of training-time RTC as overlap-aware token-wise conditioning rather than binary prefix masking;
    \item a prior-guided flow-matching objective that preserves the fast deployment profile of training-time RTC;
    \item a full-data benchmark study on released RTC assets for the 12 large Kinetix levels, using inference-time hard and soft RTC only as comparison baselines;
    \item an ablation study over delay-scaled, fixed, and offset-style soft-conditioning windows, exposing a controllable continuity-performance frontier;
    \item efficiency, qualitative, and preliminary real-robot analyses that clarify where training-time soft conditioning is practically useful.
\end{enumerate}

\section{Related Work}

\textbf{Action chunking for robot imitation.}
Chunked action prediction has become an important design pattern for high-frequency control. Action Chunking with Transformers (ACT) uses sequence prediction to improve precision and temporal consistency in fine-grained manipulation \cite{zhao2023act}. Behavior Transformer (BeT) and its vector-quantized extension VQ-BeT show that chunked transformer policies can also model highly multimodal continuous behavior through latent or discretized action tokens \cite{shafiullah2022bet,lee2024vqbet}. BAKU further illustrates how action chunking has become a standard ingredient inside stronger multi-task transformer policy stacks \cite{haldar2024baku}. Diffusion Policy similarly adopts receding-horizon action generation and demonstrates that generative action models can be strong policies for robotics \cite{chi2023diffusion}. Our work is aligned with this family of methods, but focuses on the latency problem that arises once chunked policies are deployed in real time.

\textbf{Flow-based and diffusion-based policy learning.}
Modern robot policies increasingly use generative objectives that denoise action sequences rather than predict them autoregressively. Flow Matching provides a simulation-free way to train continuous normalizing flows by regressing vector fields along probability paths \cite{lipman2022flow}. Consistency Policy is another recent route to low-latency generative control, distilling diffusion policies into much faster consistency models \cite{prasad2024consistency}. RTC explicitly targets chunked flow policies \cite{black2025rtc}, and our experiments use flow-matching policies in that setting. Soft RTC does not alter the underlying policy architecture or generative target; it changes how different tokens are corrupted and weighted so that overlap tokens learn to denoise from prior-informed states rather than from scratch.

\textbf{Real-time chunking under delay.}
The closest prior work is training-time RTC. Inference-time RTC introduces delayed asynchronous execution for chunked diffusion and flow policies by freezing committed actions and inpainting the overlap region at deployment time \cite{black2025rtc}. Training-time RTC later showed that much of the real-time execution benefit can be obtained by simulating delay and conditioning on the action prefix during training \cite{black2025train}. Soft RTC is a direct generalization of this training-time formulation: it replaces the hard prefix indicator with continuous token-wise action-prior weights. The inference-time RTC variants are therefore not algorithmic ingredients of Soft RTC; they are included as baselines because they address the same delayed-execution problem with deployment-time guidance.

\textbf{Concurrent latency-aware chunk correction.}
Recent work has started to address latency-induced chunk mismatch beyond the original RTC formulation. REMAC learns masked corrective updates on top of pretrained chunking policies and explicitly targets intra-chunk inconsistency under asynchronous execution \cite{wang2026remac}. A2C2 adds a lightweight real-time correction head for VLA action chunks and is designed to be orthogonal to RTC-style asynchronous execution \cite{sendai2025a2c2}. These methods are adjacent but conceptually different from our setting: they introduce additional corrective modules at deployment time, whereas Soft RTC changes the training-time conditioning structure of the base flow policy itself.

\textbf{Benchmark setting.}
We evaluate on the Kinetix simulator, an open-ended JAX environment family designed for diverse physics-based control tasks \cite{matthews2024kinetix}. This benchmark is suitable for RTC because its released large-task subset emphasizes dynamic execution, reactivity, and sensitivity to chunk-boundary artifacts.

\section{Problem Setting}

We consider a chunked policy that predicts an action chunk $\mathbf{A}_t=[a_t,\ldots,a_{t+H-1}]$ from observation $o_t$, where $H$ is the chunk horizon. At rollout time, only the first $s \leq H$ actions are executed before replanning, so consecutive chunks overlap over $o = H-s$ future steps.

If policy inference takes $d$ controller timesteps, then the first $d$ actions of the newly generated chunk must still come from the previous chunk. Hard training-time RTC encodes only this \emph{committed prefix}. Soft RTC introduces a second quantity, an \emph{effective conditioning endpoint} $e(d)$, which determines how far the soft-conditioning window extends into the chunk. In general, $e(d)$ is a design parameter and need not be identical to the exact physical overlap $o$.

This induces three semantic regions:
\begin{itemize}
    \item \textbf{committed prefix}: indices $j < d$;
    \item \textbf{soft-conditioning window}: indices $d \le j < e(d)$;
    \item \textbf{free tail}: indices $j \ge e(d)$.
\end{itemize}

When $e(d)=d$, the soft-conditioning window disappears and only the committed prefix remains. When $d=0$ and $e(d)=0$, the entire chunk is unconstrained and the method reduces exactly to the ordinary no-RTC training objective.

\section{Method}

\subsection{Action-Prior Denoising During Training}

Let $\epsilon \sim \mathcal{N}(0,I)$ be chunk-shaped noise and $\tau \sim \mathrm{Unif}[0,1]$ be the flow time. Standard flow matching trains $v_\theta$ to map the noisy interpolation between $\epsilon$ and $\mathbf{A}_t$ toward $\mathbf{A}_t-\epsilon$ \cite{lipman2022flow}. Training-time RTC modifies only the per-token conditioning level. Hard training-time RTC uses the binary mask $\mathbf{1}[j<d]$; Soft RTC replaces it with weights $\omega_j \in [0,1]$.

This gives a prior-guided interpretation of the training objective. For a soft-overlap token, the target action itself acts as a supervised proxy for the action prior that will later be supplied by the aligned previous chunk at deployment. Instead of learning to denoise this token from a fully unconstrained noisy state, the model learns to continue denoising from an intermediate state that already contains partial action information. The amount of prior information is controlled by $\omega_j$.

The delay-scaled endpoint and the offset endpoint used in the ablation are
\[
e(d) = \min\!\left(\left\lceil \lambda d \right\rceil, h_{\max}\right),
\qquad
e_{\mathrm{off}}(d) = \min(d+L, h_{\max}).
\]
Here $h_{\max}$ is the maximum soft-conditioning horizon and $\lambda$ is a delay multiplier. We define $e(0)=e_{\mathrm{off}}(0)=0$, so the soft-conditioning window vanishes at zero delay. The offset rule isolates the absolute soft-region length $L$ after the hard committed prefix, and makes $L{=}0$ exactly identical to Hard RTC. For a nonempty soft window, $e(d)>d$, the token weights are
\[
\omega_j =
\begin{cases}
1, & j < d, \\
g\!\left(\frac{j-d}{e(d)-d}\right), & d \le j < e(d), \\
0, & j \ge e(d),
\end{cases}
\]
with the middle case omitted when $e(d)=d$ and all weights set to zero when $d=e(d)=0$. Here $g$ is a monotone decreasing schedule. In our main experiment, $g$ is linear, $\lambda=2$, and $h_{\max}=5$. Intuitively, early overlap actions are almost fixed, later overlap actions are only partially constrained, and postfix actions remain free.

\begin{figure}[!t]
    \centering
    \includegraphics[width=0.98\columnwidth]{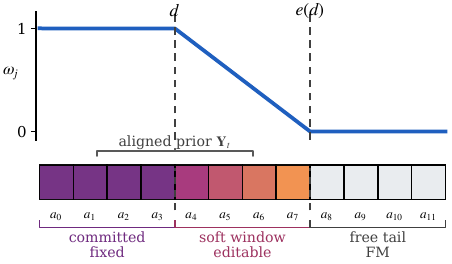}
    \caption{Soft RTC token weights for a nonzero delay. Tokens before $d$ receive full prior weight, tokens from $d$ to $e(d)$ denoise from a gradually relaxed action prior, and tokens after $e(d)$ follow ordinary flow matching.}
    \label{fig:soft-weights}
\end{figure}

The same weights define both the corrupted input and the flow-matching objective:
\[
\begin{aligned}
\tau_j^\omega &= \omega_j+(1-\omega_j)\tau,\\
x^\omega[j] &= \tau_j^\omega\mathbf{A}_t[j]+(1-\tau_j^\omega)\epsilon[j],\\
\mathcal{L}_{\omega}(\theta)
&=\mathbb{E}_{o_t,\mathbf{A}_t,d,\tau,\epsilon}
\left[
\frac{\sum_j(1-\omega_j)\ell_j^\omega(\theta)}
{\sum_j(1-\omega_j)+\varepsilon}
\right],\\
\ell_j^\omega(\theta)
&=
\left\|v_\theta(o_t,x^\omega,\tau^\omega)_j
-(\mathbf{A}_t[j]-\epsilon[j])\right\|_2^2 .
\end{aligned}
\]
This objective masks fully clamped tokens while training editable tokens with the usual flow-matching target. It covers no RTC ($\omega_j=0$), Hard RTC ($\omega_j=\mathbf{1}[j<d]$), and Soft RTC. Hard RTC is also recovered by the \texttt{zeros} soft schedule in our implementation.

For $0<\omega_j<1$, $x^\omega[j]$ is neither pure noise nor a fully clamped action: it is a prior-informed denoising state. Increasing $\omega_j$ moves the denoising problem closer to the previous action plan, while the factor $(1-\omega_j)$ keeps the token trainable and editable. This is the mechanism by which Soft RTC encourages smoother chunk transitions without turning the entire overlap into a hard constraint.

\subsection{Inference Rule}

At inference time, the previous chunk provides a reference sequence $\mathbf{Y}_t$ in the current chunk frame. Training-time RTC keeps the same explicit Euler solver as the base flow policy, but changes the denoiser inputs before each solver step. Let $t_k=k/T$. Soft RTC computes weights $\omega_j$ from the realized delay and iterates
\[
\begin{aligned}
\tilde{x}^{k}[j] &= \omega_j\mathbf{Y}_t[j]+(1-\omega_j)x^k[j],\\
\tilde{t}^{k}_j &= \omega_j+(1-\omega_j)t_k,\\
x^{k+1} &= \tilde{x}^k+\frac{1}{T}v_\theta(o_t,\tilde{x}^{k},\tilde{t}^{k}),
\end{aligned}
\]
starting from $x^0 \sim \mathcal{N}(0,I)$. Hard training-time RTC is the same rule with binary weights, so Soft RTC is obtained by making those weights continuous. The resulting deployment rule remains lightweight and adds only token-wise prior blending to the ordinary forward solver.

Although the solver state is initialized from Gaussian noise, soft-overlap tokens are not denoised from noise alone. At every solver step, $\mathbf{Y}_t$ injects an action prior into the denoiser input and also moves the token-wise flow time closer to the denoised endpoint. Thus the policy can preserve continuity with the previous chunk while still adapting the editable part of the new chunk to the latest observation.

\begin{figure}[!ht]
    \centering
    \fbox{
    \begin{minipage}{0.94\columnwidth}
    \small
    \textbf{Algorithm 1: Soft RTC Inference}\\[2pt]
    \textbf{Input:} observation $o_t$, previous chunk $\mathbf{Y}_t$, realized delay $d$, flow steps $T$\\
    1. Compute endpoint $e(d)$ and token weights $\omega_j$.\\
    2. Sample $x^0 \sim \mathcal{N}(0,I)$.\\
    3. For $k=0,\ldots,T-1$:\\
    \hspace*{1em}a. Set $t_k=k/T$.\\
    \hspace*{1em}b. Blend the state:
    $\tilde{x}^k[j]=\omega_j\mathbf{Y}_t[j]+(1-\omega_j)x^k[j]$.\\
    \hspace*{1em}c. Blend the flow time:
    $\tilde{t}^k_j=\omega_j+(1-\omega_j)t_k$.\\
    \hspace*{1em}d. Update
    $x^{k+1}=\tilde{x}^k+\frac{1}{T}v_\theta(o_t,\tilde{x}^k,\tilde{t}^k)$.\\
    4. Return $x^T$ as the new action chunk.
    \end{minipage}}
    \caption{Soft RTC inference uses the ordinary forward flow solver. The main additional operation is lightweight token-wise prior blending with the aligned previous chunk.}
    \label{fig:algorithm}
\end{figure}

\begin{figure*}[!t]
    \centering
    \subfloat[Overall solve versus high-delay jerk.]{
        \includegraphics[width=0.43\textwidth]{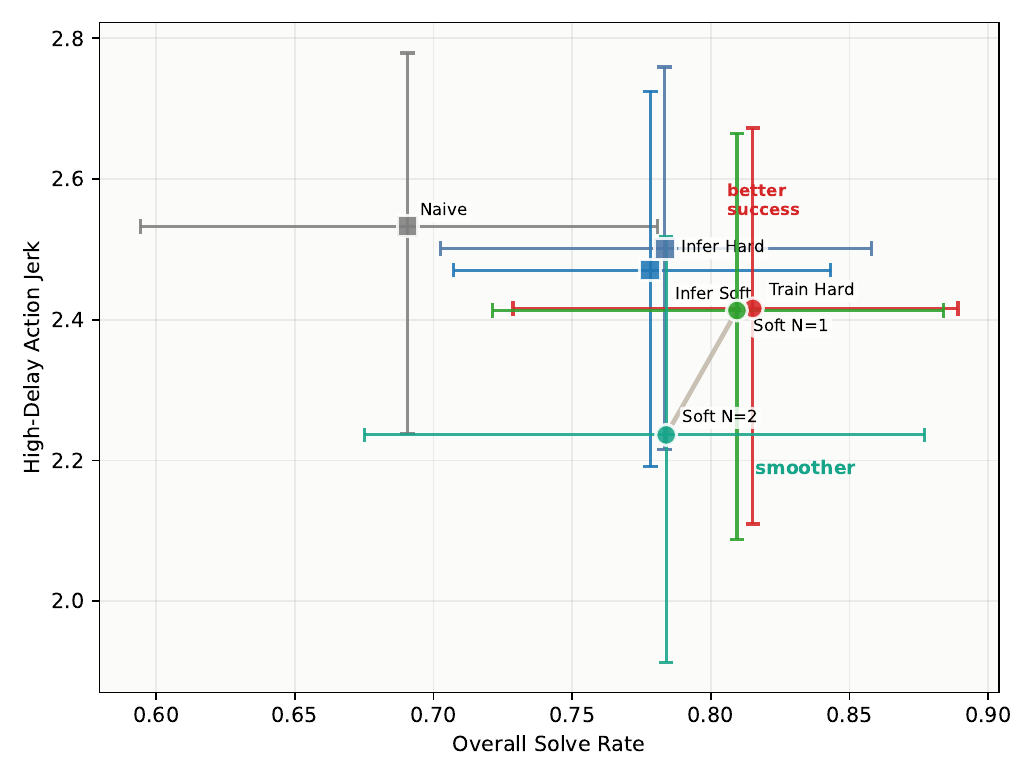}
    }
    \hfill
    \subfloat[Steady-state runtime after JIT warm-up.]{
        \includegraphics[width=0.53\textwidth]{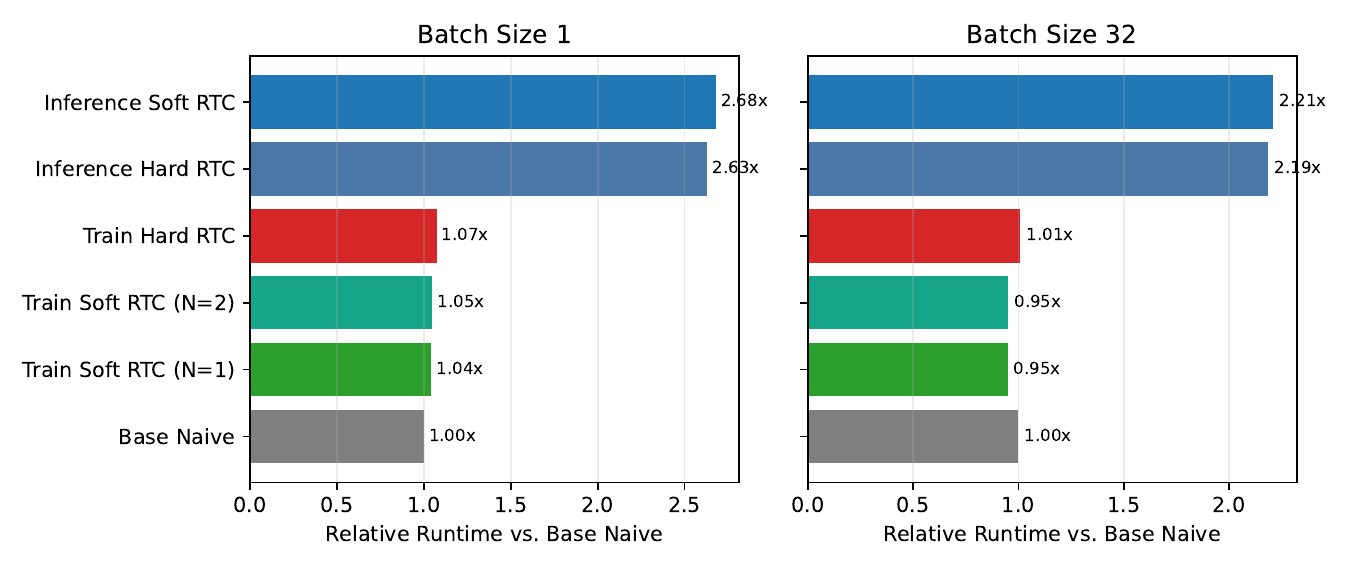}
    }
    \caption{Main practical frontier. Train Soft RTC ($N{=}1$) stays very close to Train Hard RTC on task success while preserving the same runtime profile. Train Soft RTC ($N{=}2$) moves to a lower-jerk high-delay regime while retaining the fast deployment profile of training-time RTC.}
    \label{fig:frontier-timing}
\end{figure*}

\section{Experimental Setup}

\textbf{Assets.}
We use the public RTC release described in the repository README: full-data expert demonstrations from \texttt{expert/data} and released base flow-matching checkpoints from \texttt{bc/24} \cite{black2025rtc,black2025train}.

\textbf{Tasks.}
We evaluate on all 12 released large Kinetix levels from the RTC benchmark \cite{black2025rtc,matthews2024kinetix}.

\textbf{Methods.}
We compare training-time and inference-time RTC variants separately:
\begin{itemize}
    \item \textbf{Base Naive}: released base checkpoint with naive asynchronous chunk execution;
    \item \textbf{Inference Hard RTC}: released base checkpoint with inference-time hard prefix masking;
    \item \textbf{Inference Soft RTC}: released base checkpoint with inference-time soft overlap guidance;
    \item \textbf{Train Hard RTC}: one-epoch fine-tune with simulated delay and zero schedule;
    \item \textbf{Train Soft RTC}: one-epoch fine-tune with simulated delay and delay-scaled soft-conditioning; we report both $N{=}1$ and $N{=}2$ operating points.
\end{itemize}

\textbf{Protocol.}
Both training-time variants are initialized from the released base checkpoint for the corresponding task and fine-tuned for one epoch on the released full-data dataset. Our main soft model uses delay-scaled conditioning with $h_{\max}=5$, $\lambda=2$, and a linear schedule. For Train Hard RTC and Train Soft RTC ($N{=}2$), we run three independent fine-tuning seeds; the main table reports selected operating points for the success--smoothness--runtime frontier rather than seed-averaged superiority claims. The window sweep additionally evaluates $\lambda \in \{1,2,3,4\}$, fixed windows of length 3 and 5, and an offset family $L \in \{0,1,\dots,6\}$ with $e_{\mathrm{off}}(d)=\min(d+L,8)$ and $L{=}0$ anchored to Hard RTC. The $N{=}1$ operating point reported in the main comparison comes from the delay-scaled sweep; the offset sweep is reported as a separate length-control ablation. Evaluation uses execution horizon $s=4$, inference delays $d \in \{0,1,2,3,4\}$, five flow steps, and 32 rollout episodes per level-delay-method tuple.

\textbf{Metrics.}
We report task return and solve rate, along with two continuity metrics measured on the executed action sequence:
\begin{itemize}
    \item \textbf{action delta}: mean $\ell_2$ norm of consecutive action differences;
    \item \textbf{action jerk}: mean $\ell_2$ norm of the second finite difference.
\end{itemize}
Lower continuity metrics indicate smoother cross-chunk execution.

\textbf{Efficiency benchmark.}
To quantify deployment cost, we additionally benchmark steady-state chunk-generation latency after JIT warm-up for batch sizes 1 and 32. We measure the wall-clock time of a single action-chunk generation call for each method, including inference-time guidance when applicable.

\textbf{Real-robot validation.}
We further run a preliminary single-arm real-robot validation using the LeRobot/OpenPI runtime on a four-object sorting task. The natural-language prompt is: ``Place 2 large blue metal parts $\rightarrow$ blue box, one by one. Then place 2 small silver metal parts $\rightarrow$ gray box.'' We evaluate Base, Train Hard RTC, and Train Soft RTC for 10 physical trials each. After each rollout, an operator annotates success, task completion, object completion, safety, and whether the trial is usable for paper statistics. The runtime logger automatically records commanded joint actions, observed joint states, policy-call latency, control-loop timing, chunk switches, and finite-difference smoothness metrics.

\section{Results}

\subsection{Main Comparison and Efficiency}

Table~\ref{tab:main} and Figure~\ref{fig:frontier-timing} summarize the main practical picture. We use Train Hard RTC seed 0 as a task-performance-oriented hard operating point and Train Soft RTC ($N{=}2$) seed 1 as a smoothness-oriented soft operating point. These checkpoints are used to illustrate the frontier, so the comparison should be read as operating-point evidence rather than as a seed-averaged ranking. Several points are immediate.

First, soft overlap conditioning is best understood as a \emph{family} rather than a single point. Train Soft RTC ($N{=}1$) is the performance-preserving operating point: its overall solve rate is $0.809$, only $0.6$ percentage points below Train Hard RTC's $0.815$, while its runtime remains essentially identical to naive execution. Train Soft RTC ($N{=}2$) is the smoothness-oriented operating point: its overall solve rate drops to $0.781$, but its high-delay solve rate still exceeds both inference-time baselines and its high-delay jerk is the lowest among the methods in Table~\ref{tab:main}.

Second, once runtime is accounted for, the training-time methods occupy a more favorable practical frontier. Inference Hard RTC and Inference Soft RTC require $2.63\times$ and $2.68\times$ the naive runtime at batch size 1, and $2.19\times$ and $2.21\times$ at batch size 32. In contrast, Train Hard RTC, Train Soft RTC ($N{=}1$), and Train Soft RTC ($N{=}2$) remain within $1.07\times$, $1.04\times$, and $1.05\times$ of naive runtime at batch size 1, and are effectively runtime-neutral at batch size 32.

Bootstrap intervals over levels reinforce the right interpretation of these points. For overall solve, Train Hard RTC and Train Soft RTC ($N{=}1$) have heavily overlapping $95\%$ bootstrap intervals, $[0.730, 0.889]$ and $[0.722, 0.885]$, so we do not claim a statistically clean success advantage for either method. Instead, the evidence supports a practical frontier view: the short-window soft model tracks the hard model closely on task success, while the medium-window model moves toward a smoother high-delay controller.

\begin{table*}[t]
    \centering
    \small
    \caption{Representative simulator comparison. Higher is better for return and solve; lower is better for jerk and runtime.}
    \label{tab:main}
    \begin{tabular}{lcccccc}
    \toprule
    & \multicolumn{2}{c}{Overall} & \multicolumn{2}{c}{High Delay ($d \in \{3,4\}$)} & \multicolumn{2}{c}{Runtime / Naive} \\
    Method & Return & Solve & Solve & Jerk $\downarrow$ & Batch 1 & Batch 32 \\
    \midrule
    Base Naive & 0.662 & 0.691 & 0.531 & 2.533 & 1.00x & 1.00x \\
    Inference Hard RTC & 0.777 & 0.783 & 0.693 & 2.501 & 2.63x & 2.19x \\
    Inference Soft RTC & 0.773 & 0.778 & 0.678 & 2.471 & 2.68x & 2.21x \\
    Train Hard RTC & \textbf{0.838} & \textbf{0.815} & \textbf{0.732} & 2.416 & 1.07x & 1.01x \\
    Train Soft RTC ($N{=}1$) & 0.832 & 0.809 & 0.724 & 2.413 & 1.04x & 0.95x \\
    Train Soft RTC ($N{=}2$) & 0.792 & 0.781 & 0.702 & \textbf{2.185} & 1.05x & 0.95x \\
    \bottomrule
    \end{tabular}
\end{table*}

The $N{=}2$ operating point suggests that prior-informed corruption changes the learned success--smoothness trade-off beyond the hard-mask baseline. Relative to Train Hard RTC, it reduces high-delay action delta from $1.403$ to $1.276$ and high-delay action jerk from $2.416$ to $2.185$, corresponding to $9.1\%$ and $9.6\%$ reductions, while keeping near-naive runtime. Relative to the inference-time baselines, it is faster and has lower high-delay jerk, and it also attains higher high-delay solve than either Inference Hard RTC ($0.693$) or Inference Soft RTC ($0.678$).

\subsection{Preliminary Real-Robot Validation}

Table~\ref{tab:real-robot} summarizes the physical single-arm sorting trials. The main finding is consistent with the simulator story but not identical to it. Both RTC-trained policies improve real-robot task outcome over the base policy: Base succeeds in 6 of 10 trials and completes 33 of 40 target placements, Train Hard RTC succeeds in 9 of 10 trials and completes 39 of 40 placements, and Train Soft RTC succeeds in 8 of 10 trials and completes 37 of 40 placements. Qualitatively, the base policy sometimes exhibits back-and-forth oscillation and stalls around a local region before completing the next manipulation phase; these episodes increase task duration and can end in missed or incomplete placements. This is reflected in object-level throughput: Base completes $3.18$ objects/min, while Train Hard RTC and Train Soft RTC reach $5.07$ and $5.21$ objects/min.

The motion metrics show a complementary pattern. Relative to Base, Train Soft RTC reduces the mean commanded second and third finite differences by approximately $49.7\%$ and $52.3\%$, and it produces the smallest chunk-boundary action jump. Train Hard RTC also reduces the same second and third finite differences by $37.1\%$ and $39.7\%$, but it is less smooth than Train Soft RTC in this hardware run. Thus the present real-robot evidence supports a cautious interpretation: training-time RTC improves both completion and continuity over the base policy in this setting, while Hard RTC gives the best physical completion and Soft RTC gives the lowest commanded-action finite-difference metrics and highest completed-object throughput among the tested policies.

The latency columns should also be interpreted as implementation measurements rather than theoretical RTC overhead. The base policy averages $60.6$ ms per policy call, while the two RTC-trained checkpoints average roughly $87$ ms in the current OpenPI deployment. Since these are training-time RTC policies, the measured difference reflects the deployed checkpoint/server configuration rather than an inference-time correction cost. We do not include inference-time RTC in the hardware table because those trials are not yet available.

\input{tables/real_robot_summary}

\subsection{Delay-Scaled Window Sweep}

Figure~\ref{fig:sweeps}(a) shows the soft-window sweep. Two settings occupy the most useful parts of the frontier. The delay-scaled $N{=}1$ rule is the most performance-preserving soft variant, while the delay-scaled $N{=}2$ rule is the most favorable smoothness-oriented soft variant in our sweep. Larger multipliers continue to lower jerk slightly but do so by moving sharply left on solve rate. Fixed windows are also less favorable defaults: a long fixed window ($h{=}5$) is too aggressive, while a shorter fixed window ($h{=}3$) remains behind the delay-aware alternatives.

This sweep is important because it changes the interpretation of the method. The question is not whether a single soft-conditioning rule beats hard RTC everywhere. Instead, soft overlap conditioning exposes a training-time smoothness--performance frontier whose location can be controlled by a single delay-scaling hyperparameter.

\subsection{Offset-Length Sweep}

\begin{figure*}[t]
    \centering
    \subfloat[Delay-scaled and fixed soft windows.]{
        \includegraphics[width=0.47\textwidth]{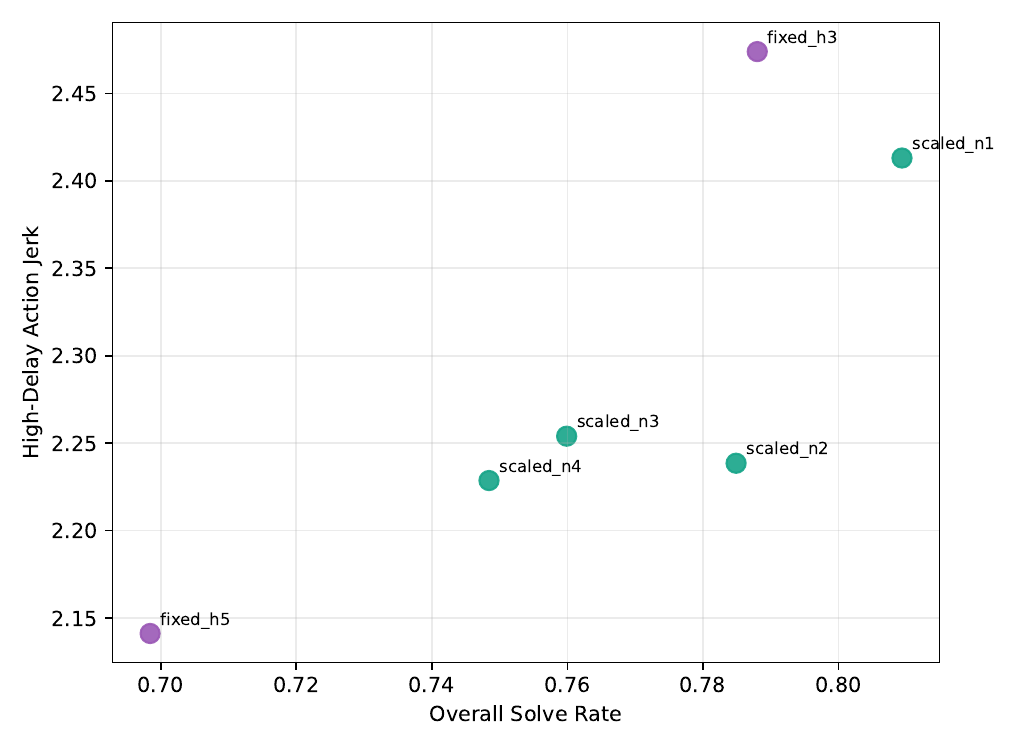}
    }
    \hfill
    \subfloat[Offset-length rule $e_{\mathrm{off}}(d)=\min(d+L,8)$.]{
        \includegraphics[width=0.49\textwidth]{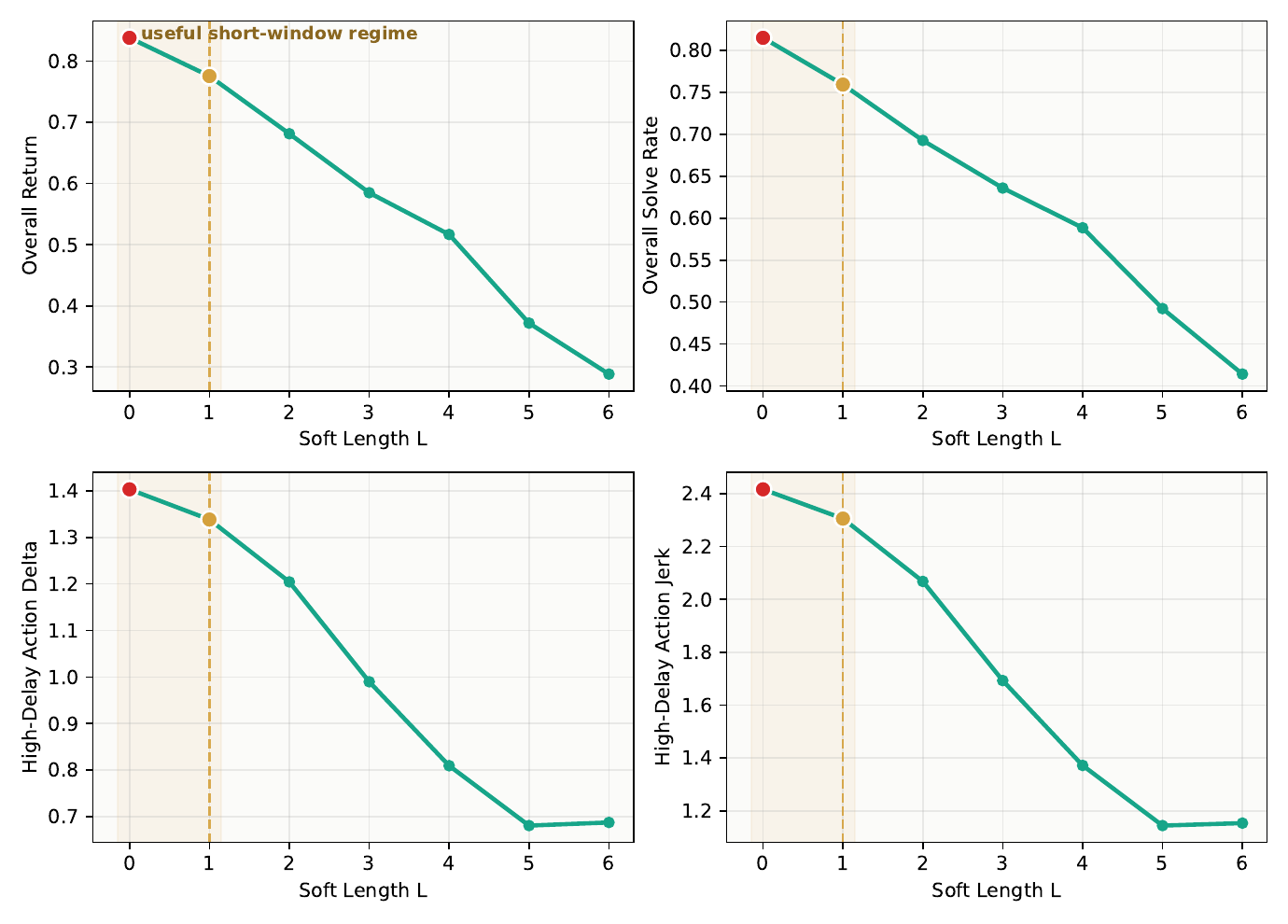}
    }
    \caption{Soft-window ablations. Delay-aware scaling exposes two useful operating points, while the offset sweep shows that longer editable overlap windows improve continuity at the cost of rapid success degradation.}
    \label{fig:sweeps}
\end{figure*}

The delay-scaled sweep above changes the endpoint together with delay. Figure~\ref{fig:sweeps}(b) complements it with a stricter ablation that isolates the \emph{absolute} soft-region length after the hard committed prefix. Here $L{=}0$ recovers Hard RTC exactly, and increasing $L$ appends a longer editable-but-biased region to every delayed chunk.

This experiment confirms the expected monotonic trade-off, but it also shows that the practical regime is short. As $L$ increases, high-delay action delta and jerk fall almost monotonically, yet return and solve degrade quickly. Hard RTC ($L{=}0$) remains the strongest task-performance point overall. The best offset-style soft variant is $L{=}1$: relative to Hard RTC, it lowers high-delay delta and jerk by $4.6\%$ and $4.6\%$, while overall solve drops from $0.815$ to $0.759$. Longer windows rapidly over-regularize the policy. At $L{=}2$, high-delay delta and jerk improve by $14.2\%$ and $14.4\%$, but high-delay solve already falls from $0.732$ to $0.486$; by $L{=}5$ and $L{=}6$, high-delay solve is only $0.124$ and $0.115$.

This offset ablation is useful for two reasons. First, it validates the semantics of the implementation: the $L{=}0$ endpoint collapses to Hard RTC, as intended. Second, it sharpens the design lesson behind Soft RTC. The soft-conditioning window is a genuine control knob, but longer is not better. The useful part of the frontier lies in short windows, which explains why the delay-aware $N{=}1$ and $N{=}2$ rules are more favorable defaults than large fixed or offset-style windows.

\section{Discussion}

The main conclusion is not that one soft window universally replaces hard RTC. The more useful conclusion is that action-prior denoising changes the learned controller and that this change can be tuned. Train Hard RTC remains the strongest single point if the only objective is maximizing average task success. But soft overlap conditioning exposes at least two practically useful alternatives: a short window that almost preserves the hard-RTC controller, and a medium window that substantially improves delayed smoothness while retaining the runtime profile of training-time RTC. The new offset-length ablation reinforces this interpretation by showing a sharp frontier: once the editable overlap grows beyond a very short region, the learned controller becomes visibly over-constrained.

This distinction matters because RTC practice implicitly mixes three goals: preserving success under latency, preserving motion quality under latency, and preserving wall-clock deployability. The real-robot observations make this practical rather than cosmetic: a chunked base policy may still produce individually plausible action chunks, yet latency and jitter can cause stale-action execution and inconsistent chunk handoff, which appear as oscillation, local stalling, and delayed or failed manipulation. Inference-time hard and soft RTC can help with the first two goals, but they are the slowest methods in our runtime benchmark. The training-time soft family separates these axes more cleanly. Short windows stay near the hard-RTC success point. Medium windows move toward smoother delayed behavior. Both retain the fast deployment behavior of training-time RTC.

These results should be read with a few caveats: the training variants are one-epoch fine-tunes from released checkpoints, the soft-window sweep covers a limited set of endpoint rules, and the hardware study is still preliminary. Broader real-robot evaluation and adaptive endpoint selection remain future work.

\section{Conclusion}

We presented Soft RTC, an action-prior denoising approach for training-time real-time chunking with flow-based action policies. The method directly generalizes hard training-time RTC by replacing binary prefix conditioning with token-wise overlap weights, so editable overlap tokens learn to denoise from partially prior-informed states instead of from scratch. It preserves the fast deployment profile of training-time RTC and exactly recovers ordinary no-RTC behavior when delay is zero. In a full-data Kinetix study where inference-time RTC variants serve only as comparison baselines, we find that soft overlap conditioning is most useful when viewed as a family of operating points. A short delay-scaled window nearly matches hard training-time RTC on overall task success, while a medium delay-scaled window improves high-delay smoothness and compares favorably with inference-time baselines on the success--smoothness--runtime frontier. A complementary offset-length sweep shows that this frontier is steep: short soft windows are useful, but long ones quickly over-constrain delayed control. A preliminary single-arm hardware study further suggests that training-time RTC can improve physical task completion and command continuity over the base policy, with Hard RTC giving the highest completion and Soft RTC giving the lowest commanded-action finite-difference metrics on the tested sorting task. The key design question is how the effective soft-conditioning window should scale with delay and task demands.

\end{document}

%% file: tables/real_robot_summary.tex
\begin{table*}[t]
    \centering
    \scriptsize
    \caption{Real-robot single-arm sorting evaluation with 10 physical trials per method. Throughput is completed objects per minute; $D_1$, $D_2$, and $D_3$ are commanded-action finite differences in action units; Bound. is the chunk-boundary action jump; Lat. is policy-call latency in milliseconds. Values with $\pm$ report standard error.}
    \label{tab:real-robot}
    \setlength{\tabcolsep}{3pt}
    \begin{tabular}{@{}lccccccccc@{}}
    \toprule
    Method & Succ. & Compl. & Obj. & Thr. & $D_1\downarrow$ & $D_2\downarrow$ & $D_3\downarrow$ & Bound. $\downarrow$ & Lat. \\
    \midrule
    Base & 6/10 & 0.825 $\pm$ 0.084 & 33/40 & 3.18 $\pm$ 0.37 & 0.0066 $\pm$ 0.0001 & 0.0047 $\pm$ 0.0001 & 0.0088 $\pm$ 0.0003 & 0.0520 $\pm$ 0.0074 & 60.6 $\pm$ 0.1 \\
    Train Hard RTC & 9/10 & 0.975 $\pm$ 0.025 & 39/40 & 5.07 $\pm$ 0.37 & 0.0062 $\pm$ 0.0001 & 0.0030 $\pm$ 0.0001 & 0.0053 $\pm$ 0.0001 & 0.0285 $\pm$ 0.0052 & 87.0 $\pm$ 0.2 \\
    Train Soft RTC & 8/10 & 0.925 $\pm$ 0.053 & 37/40 & 5.21 $\pm$ 0.44 & 0.0061 $\pm$ 0.0001 & 0.0024 $\pm$ 0.0000 & 0.0042 $\pm$ 0.0001 & 0.0118 $\pm$ 0.0027 & 87.5 $\pm$ 0.2 \\
    \bottomrule
    \end{tabular}
\end{table*}

%% file: main.bbl
\begin{thebibliography}{99}

\bibitem{black2025rtc}
Kevin Black, Manuel Y. Galliker, and Sergey Levine.
\newblock Real-Time Execution of Action Chunking Flow Policies.
\newblock \emph{arXiv preprint arXiv:2506.07339}, 2025.

\bibitem{black2025train}
Kevin Black, Allen Z. Ren, Michael Equi, and Sergey Levine.
\newblock Training-Time Action Conditioning for Efficient Real-Time Chunking.
\newblock \emph{arXiv preprint arXiv:2512.05964}, 2025.

\bibitem{chi2023diffusion}
Cheng Chi, Zhenjia Xu, Siyuan Feng, Eric Cousineau, Yilun Du, Benjamin Burchfiel, Russ Tedrake, and Shuran Song.
\newblock Diffusion Policy: Visuomotor Policy Learning via Action Diffusion.
\newblock \emph{arXiv preprint arXiv:2303.04137}, 2023.

\bibitem{lipman2022flow}
Yaron Lipman, Ricky T. Q. Chen, Heli Ben-Hamu, Maximilian Nickel, and Matt Le.
\newblock Flow Matching for Generative Modeling.
\newblock \emph{arXiv preprint arXiv:2210.02747}, 2022.

\bibitem{matthews2024kinetix}
Michael Matthews, Michael Beukman, Chris Lu, and Jakob Foerster.
\newblock Kinetix: Investigating the Training of General Agents through Open-Ended Physics-Based Control Tasks.
\newblock \emph{arXiv preprint arXiv:2410.23208}, 2024.

\bibitem{haldar2024baku}
Siddhant Haldar, Zhuoran Peng, and Lerrel Pinto.
\newblock BAKU: An Efficient Transformer for Multi-Task Policy Learning.
\newblock \emph{arXiv preprint arXiv:2406.07539}, 2024.

\bibitem{lee2024vqbet}
Seungjae Lee, Yibin Wang, Haritheja Etukuru, H. Jin Kim, Nur Muhammad Mahi Shafiullah, and Lerrel Pinto.
\newblock Behavior Generation with Latent Actions.
\newblock \emph{arXiv preprint arXiv:2403.03181}, 2024.

\bibitem{prasad2024consistency}
Aaditya Prasad, Kevin Lin, Jimmy Wu, Linqi Zhou, and Jeannette Bohg.
\newblock Consistency Policy: Accelerated Visuomotor Policies via Consistency Distillation.
\newblock \emph{arXiv preprint arXiv:2405.07503}, 2024.

\bibitem{shafiullah2022bet}
Nur Muhammad Mahi Shafiullah, Zichen Jeff Cui, Ariuntuya Altanzaya, and Lerrel Pinto.
\newblock Behavior Transformers: Cloning $k$ modes with one stone.
\newblock \emph{arXiv preprint arXiv:2206.11251}, 2022.

\bibitem{sendai2025a2c2}
Kohei Sendai, Maxime Alvarez, Tatsuya Matsushima, Yutaka Matsuo, and Yusuke Iwasawa.
\newblock Leave No Observation Behind: Real-time Correction for VLA Action Chunks.
\newblock \emph{arXiv preprint arXiv:2509.23224}, 2025.

\bibitem{wang2026remac}
Haoxuan Wang, Gengyu Zhang, Yan Yan, Yuzhang Shang, Ramana Rao Kompella, and Gaowen Liu.
\newblock Real-Time Robot Execution with Masked Action Chunking.
\newblock \emph{arXiv preprint arXiv:2601.20130}, 2026.

\bibitem{zhao2023act}
Tony Z. Zhao, Vikash Kumar, Sergey Levine, and Chelsea Finn.
\newblock Learning Fine-Grained Bimanual Manipulation with Low-Cost Hardware.
\newblock \emph{arXiv preprint arXiv:2304.13705}, 2023.

\end{thebibliography}
